%% file: main.tex
\documentclass[10pt, a4paper]{article}
\usepackage{lrec2022} % this is the new LREC2022 Style
\usepackage{multibib}
\usepackage{microtype}
\newcites{languageresource}{Language Resources}
\usepackage{graphicx}
\usepackage{tabularx}
\usepackage{soul}

\usepackage{multirow}
\usepackage{booktabs}
\usepackage[nolist]{acronym}
\input{acronyms}
\usepackage{footmisc}

\usepackage{amssymb}% http://ctan.org/pkg/amssymb
\usepackage{pifont}% http://ctan.org/pkg/pifont
\newcommand{\cmark}{\ding{51}}%
\newcommand{\xmark}{\ding{55}}%

\usepackage[dvipsnames]{xcolor}

\usepackage{titlesec}
\titleformat{\section}{\normalfont\large\bfseries\center}{\thesection.}{1em}{}
\titleformat{\subsection}{\normalfont\SmallTitleFont\bfseries\raggedright}{\thesubsection.}{1em}{}
\titleformat{\subsubsection}{\normalfont\normalsize\bfseries\raggedright}{\thesubsubsection.}{1em}{}
\renewcommand\thesection{\arabic{section}}
\renewcommand\thesubsection{\thesection.\arabic{subsection}}
\renewcommand\thesubsubsection{\thesubsection.\arabic{subsubsection}}
\usepackage{epstopdf}
\usepackage[utf8]{inputenc}

\usepackage{hyperref}
\usepackage[capitalize]{cleveref}
\usepackage{xstring}

\usepackage{color}

\title{\vspace*{.5\baselineskip} \textbf{Changing the Representation: Examining Language Representation for Neural Sign Language Production}} 

\name{Harry Walsh, Ben Saunders, Richard Bowden} 

\address{\\ University of Surrey\\
         \{harry.walsh, b.saunders, r.bowden\}@surrey.ac.uk\\}

\abstract{
Neural \ac{slp} aims to automatically translate from spoken language sentences to sign language videos. Historically the \ac{slp} task has been broken into two steps; Firstly, translating from a spoken language sentence to a gloss sequence and secondly, producing a sign language video given a sequence of glosses. In this paper we apply Natural Language Processing techniques to the first step of the \ac{slp} pipeline. We use language models such as BERT and Word2Vec to create better sentence level embeddings, and apply several tokenization techniques, demonstrating how these improve performance on the low resource translation task of Text to Gloss. We introduce Text to HamNoSys (T2H) translation, and show the advantages of using a phonetic representation for sign language translation rather than a sign level gloss representation. Furthermore, we use HamNoSys to extract the hand shape of a sign and use this as additional supervision during training, further increasing the performance on T2H. Assembling best practise, we achieve a BLEU-4 score of 26.99 on the MineDGS dataset and 25.09 on PHOENIX14T, two new state-of-the-art baselines.  
 \\ \newline  }

\begin{document}

\maketitleabstract

\input{sections/1_intro}

\input{sections/2_related_work}

\input{sections/3_method}

\input{sections/4_eperimental}

\input{sections/5_quantitative}

\input{sections/6_qualitative}

\section{Conclusion}
In this paper, we employed a transformer to translate from spoken language sentences to a sequence of gloss or HamNoSys. We introduced T2H translation, showing the advantages of translating to HamNoSys instead of just gloss, and set baseline results for future work on mDGS.  We showed that language models can be used to improve translation performance, but using more advanced tokenization algorithms like \ac{bpe} gives a larger performance gain. Additionally, we have shown that translation can be improved by training the model to jointly predict hand shape and HamNoSys. We achieved a BLEU-4 score of 26.99 and 25.09, a new state-of-the-arts for \ac{slt} on the mDGS and \ac{ph14t} datasets.

\quad As future work, it would be interesting to create a representation, gloss++. This could combine the benefits of gloss and HamNoSys, including non-manual features as well as hand shape information, as this has been shown to be useful for translation. Furthermore, this could be beneficial for down stream tasks in the \ac{slp} pipeline.  

%-------------------------------------------------------------------------------------------------------------------
%-------------------------------------------------------------------------------------------------------------------

\section{Acknowledgements}
%We thanks Intel for supporting this research. 
We thank Adam Munder, Mariam Rahmani and Marina Lovell from OmniBridge, an Intel Venture, for supporting this project.
We also thank Thomas Hanke and University of Hamburg for use of the \ac{mdgs} data.
We also thank the SNSF Sinergia project ‘SMILE II’ (CRSII5 193686), the European Union’s Horizon2020 research project EASIER (101016982) and the EPSRC project ‘ExTOL’ (EP/R03298X/1). This work reflects only the authors view and the Commission is not responsible for any use that may be made of the information it contains.

\section{Bibliographical References}\label{reference}

\bibliographystyle{lrec2022-bib}
\bibliography{main}

\section{Language Resource References}
\label{lr:ref}
\bibliographystylelanguageresource{lrec2022-bib}
\bibliographylanguageresource{languageresource}

\end{document}

%% file: acronyms.tex
\begin{acronym}[PHOENIX14\textbf{T} ]

\acrodefplural{rnn}[RNNs]{Recurrent Neural Networks}
\acrodefplural{cnn}[CNNs]{Convolutional Neural Networks}
\acrodefplural{hmm}[HMMs]{Hidden Markov Models}
\acrodefplural{gru}[GRUs]{Gated Recurrent Units}
\acrodefplural{crf}[CRFs]{Conditional Random Fields}
\acrodefplural{gan}[GANs]{Generative Adversarial Networks}
\acrodefplural{gpu}[GPUs]{Graphic Processing Units}

\acrodefplural{mdn}[MDNs]{Mixture Density Networks}

\acro{bpe}[BPE]{Byte Pair Encoding}
\acro{bsl}[BSL]{British Sign Language}
\acro{bleu}[BLEU]{Bilingual Evaluation Understudy}
\acro{blstm}[BLSTM]{Bidirectional Long Short-Term Memory}
\acro{bslcpt}[BSLCP\textbf{T}]{BSL Corpus \textbf{T}}
\acro{cnn}[CNN]{Convolutional Neural Network}
\acro{crf}[CRF]{Conditional Random Field}
\acro{cslr}[CSLR]{Continuous Sign Language Recognition}
\acro{ctc}[CTC]{Connectionist Temporal Classification}
\acro{c4a}[C4A]{Content4All}
\acro{dl}[DL]{Deep Learning}
\acro{dgs}[DGS]{German Sign Language - Deutsche Gebärdensprache}
\acro{dsgs}[DSGS]{Swiss German Sign Language - Deutschschweizer Geb\"ardensprache}
\acro{dtw}[DTW]{Dynamic Time Warping}
\acro{fc}[FC]{Fully Connected}
\acro{ff}[FF]{Feed Forward}
\acro{gan}[GAN]{Generative Adversarial Network}
\acro{gpu}[GPU]{Graphics Processing Unit}
\acro{gru}[GRU]{Gated Recurrent Unit}
\acro{gtpt}[G2PT]{Gloss to Pose Transformer}
\acro{hmm}[HMM]{Hidden Markov Model}
\acro{hpe}[HPE]{Hand Pose Enhancer}
\acro{isl}[ISL]{Irish Sign Language}
\acro{lstm}[LSTM]{Long Short-Term Memory}
\acro{mha}[MHA]{Multi-Headed Attention}
\acro{mtc}[MTC]{Monocular Total Capture}
\acro{mse}[MSE]{Mean Squared Error}
\acro{mdn}[MDN]{Mixture Density Network}
\acro{mdgs}[mDGS]{MineDGS}
\acro{mdgst}[mDGS\textbf{T}]{meineDGS\textbf{T}}
\acro{mdgsth}[mDGS\textbf{T}-\textbf{H}]{meineDGS\textbf{T}-\textbf{HARD}}
\acro{mdgste}[mDGS\textbf{T}-\textbf{E}]{meineDGS\textbf{T}-\textbf{EASY}}
\acro{mt}[MT]{Machine Translation}

\acro{nmt}[NMT]{Neural Machine Translation}
\acro{nlp}[NLP]{Natural Language Processing}
\acro{ph12}[PHOENIX12]{RWTH-PHOENIX-Weather-2012}
\acro{ph14}[PHOENIX14]{RWTH-PHOENIX-Weather-2014}
\acro{ph14t}[PHOENIX14\textbf{T}]{RWTH-PHOENIX-Weather-2014\textbf{T}}
\acro{pof}[POF]{Part Orientation Field}
\acro{paf}[PAF]{Part Affinity Field}
\acro{pttt}[P2TT]{Pose to Text Transformer}
\acro{relu}[RELU]{Rectified Linear Units}
\acro{rnn}[RNN]{Recurrent Neural Network}
\acro{rouge}[ROUGE]{Recall-Oriented Understudy for Gisting Evaluation}
\acro{sgd}[SGD]{Stochastic Gradient Descent}
\acro{sla}[SLA]{Sign Language Assessment}
\acro{slr}[SLR]{Sign Language Recognition}
\acro{slt}[SLT]{Sign Language Translation}
\acro{slp}[SLP]{Sign Language Production}
\acro{smt}[SMT]{Statistical Machine Translation}

\acro{ttgt}[T2GT]{Text to Gloss Transformer}
\acro{ttpt}[T2PT]{Text to Pose Transformer}
\acro{ttp}[T2P]{Text to Pose}
\acro{ttg}[T2G]{Text to Gloss}
\acro{tth}[T2H]{Text to HamNoSys}
\acro{ttgth}[T2G2H]{Text to Gloss to HamNoSys}
\acro{ttgtp}[T2G2P]{Text to Gloss to Pose}
\acro{tts}[TTS]{Text to Speech}
\acro{wer}[WER]{Word Error Rate}

\end{acronym}

%% file: sections/1_intro.tex
\section{Introduction}

Sign languages are the dominant form of communication for Deaf communities, with 430 million users worldwide \cite{who_2021}. Sign languages are complex multichannel languages with their own grammatical structure and vocabulary \cite{stokoe1980sign}. For many people, sign language is their primary language, and written forms of spoken language are their secondary languages.

\quad \acf{slp} aims to bridge the gap between hearing and Deaf communities, by translating from spoken language sentences to sign language sequences. This problem has historically been broken into two steps; 1) translation from spoken language to gloss\footnote{Gloss is the written word associated with a sign \label{fngloss}} and 2) subsequent production of sign language sequences from a sequence of glosses, commonly using a graphical avatar \cite{elliott2008linguistic,efthimiou2010dicta,efthimiou2009sign} or more recently, a photo-realistic signer \cite{saunders2021anonysign,saunders2021mixed}. In this paper, we improve the \ac{slp} pipeline by focusing on the \acf{ttg} translation task of step 1.

\quad Modern deep learning is heavily dependent upon data. However, the creation of sign language datasets is both time consuming and costly, restricting their size to orders of magnitude smaller than their spoken language counterparts. State-of-the-art datasets such as \ac{ph14t}, and the newer \ac{mdgs}, contain only 8,257 and 63,912 examples respectively \citelanguageresource{KOLLER2015108,jahn:18018:sign-lang:lrec}, compared to over 15 million examples for common spoken language datasets \cite{vrandevcic2014wikidata}. Hence, sign languages can be considered as low resource languages. 

\quad In this work, we take inspiration from \ac{nlp} techniques to boost translation performance. We explore how language can be modeled using different tokenizers, more specifically \ac{bpe}, WordPiece, word and character level tokenizers. We show that finding the correct tokenizer for the task helps simplify the translation problem. 

\quad Furthermore, to help tackle our low resource language task, we explore using pre-trained language models such as BERT \cite{devlin2018bert} and Word2Vec \cite{mikolov2013distributed} to create improved sentence level embeddings. We also fuse contextual information from the embedding to increase the amount of information available to the network. We show that using models trained on large corpuses of data improves translation performance. 

\begin{figure}[ht]
\begin{center}
\includegraphics[width=0.45\textwidth]{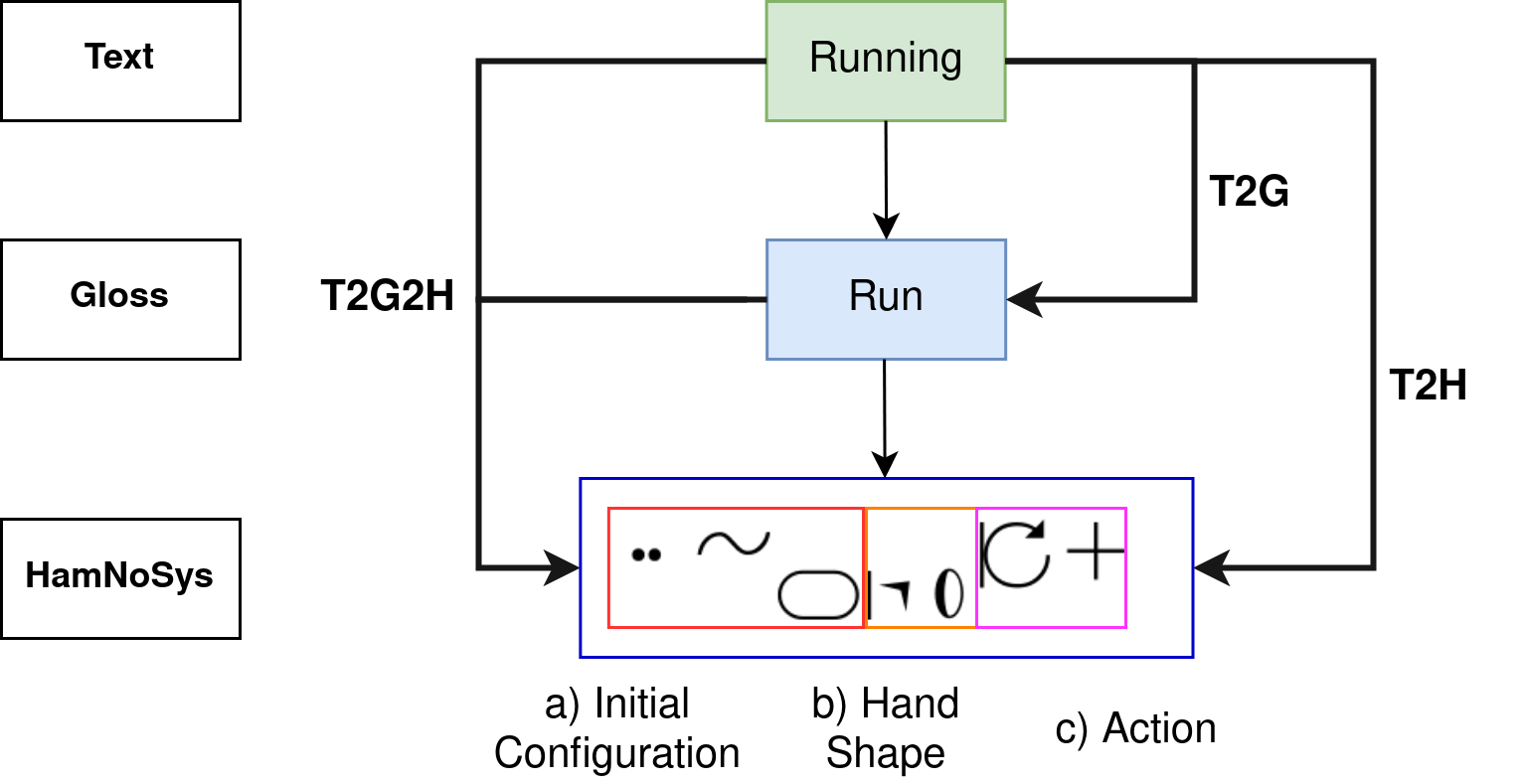} 
\caption{A graph to show the word ``running" which would be `glossed' as RUN and the associated sequence of HamNoSys, Top: Text, Middle: Gloss, Bottom: HamNoSys. HamNoSys is split into: a) it's initial configuration b) it's hand shape 3) it's action}
\label{translation}
\end{center}
\end{figure}

\quad  Previously the first step of the \ac{slp} pipeline used \ac{ttg} translation. We explore using a phonetic representation based on the Hamburg Notation System (HamNoSys) which we define as \ac{tth}. HamNoSys encodes signs using a set of symbols and can be viewed as a phonetic representation of sign language \cite{hanke2004hamnosys}. There are three main components when representing a sign in HamNoSys; a) its initial configuration b) it's hand shape and c) it's action. An example of HamNoSys can be seen in \cref{translation} along with its gloss and text counterparts.

\quad We evaluate our \ac{slp} models on both the \ac{mdgs} and \ac{ph14t} datasets, showing state-of-the-art performance on \ac{ttg} (\ac{mdgs} \& PHX) and \ac{tth} (\ac{mdgs}) tasks. We achieve a BLEU-4 score of 26.99 on \ac{mdgs}, a significant increase compared to the state-of-the-art score of 3.17 \cite{saunders2021signing}. 

\quad The rest of this paper is structured as follows; In section 2 we review the related work in the field. Section 3 presents our methodology. Section 4 shows quantitative and qualitative results. Finally, we draw conclusions in section 5 and suggest future work.     

%% file: sections/2_related_work.tex
\section{Related Work}

\textbf{Sign Language Recognition \& Translation:} Computational sign language research has been studied for over 30 years \cite{tamura1988recognition}. Research started with isolated \ac{slr} where individual signs were classified using CNNs \cite{Lecun_Gradient_based_lrn}. Recently, the field has moved to the more challenging problem of \ac{cslr}, where a continuous sign language video needs to be segmented and then classified \citelanguageresource{KOLLER2015108}. Most modern approaches to \ac{slr} and \ac{cslr} rely on deep learning, but such approaches are data hungry and therefore are limited by the size of publicly available datasets.
 
\quad The distinction between \ac{cslr} and \acf{slt} was stressed by \newcite{camgoz2018neural}. \ac{slt} aims to translate a continuous sequence of signs to spoken language sentences (Sign to Text (S2T)) or vice versa (Text to Sign (T2S)), a challenging problem due to the  changes in grammar and sequence ordering.

\textbf{\acf{slp}:} focusses on T2S, the production of a continuous sign language sequence given a spoken language input sentence. Current state-of-the-art approaches to \ac{slp} use transformer based architectures with attention \cite{stoll2018sign,saunders2020progressive}. In this paper, we tackle the \ac{slp} task of neural sign language translation, defined as \ac{ttg} or \ac{tth} translation.

\quad HamNoSys has been used before for statistical \ac{slp}, with some success \cite{kaur2014hamnosys,kaur2016hamnosys}. However, the produced motion becomes robotic and is not practical for real world applications. Note that these approaches first convert the HamNoSys to SiGML, an XML format of HamNoSys \cite{kaur2016hamnosys}.

\textbf{\acf{nmt}:} \ac{nmt} aims to generate a target sequence given a source sequence using neural networks \cite{bahdanau2014neural} and is commonly used for spoken language translations. Initial approaches used recurrence to map a hidden state to an output sequence \cite{kalchbrenner2013recurrent}, with limited performance. Encoder-decoder structures were later introduced, that map an input sequence to an embedding space \cite{wu2016google}. To address the bottleneck problem, attention was introduced to measure the affinity between sections of the input and embedding space and allow the model to focus on specific context \cite{bahdanau2014neural}. This was improved further with the introduction of the transformer \cite{vaswani2017attention} that used \ac{mha} to allow multiple projections of the learned attention. More recently, model sizes have grown with architectures introduced such as GPT-2 \cite{radford2019language} and BERT \cite{devlin2018bert}. 

Different encoding/decoding schemes have been explored. \quad \ac{bpe} was first introduced in \newcite{DBLP:journals/corr/SennrichHB15}, to create a set of tokens given a set vocabulary size. This is achieved by merging the most commonly occurring sequential characters. WordPiece, a similar tokenizer to BPE, was first introduced in \newcite{schuster2012japanese} and is commonly used when training language models such as BERT, DistilBERT and Electra. Finally, word and character level tokenizers break up a sentence based on white space and unique symbols respectively.

\textbf{Natural Language Processing:} \ac{nlp} has many applications, for example Text Simplification, Text Classification, and Speech Recognition. Recently, deep learning approaches have outperformed older statistical methods \cite{vaswani2017attention}. A successful \ac{nlp} model must understand the structure and context of language, learned via supervised or unsupervised methods. Pre-trained language models have been used to boost performance in other \ac{nlp} tasks \cite{clinchant2019use,zhu2020incorporating}, such as BERT \cite{devlin2018bert} achieving state-of-the-art performance. Zhu et al., 2020 tried to fuse the embedding of BERT into a traditional transformer architecture using attention, increasing the translation performance by approximately 2 BLEU score.

\quad Other methods have used Word2Vec to model language, this has been applied to many \ac{nlp} tasks \cite{mikolov2013distributed}. Word2Vec is designed to give meaning to a numerical representation of words. The central idea being that words with similar meaning should have a small euclidean distance between the vector representation. 

\quad In this paper, we take inspiration from these techniques to boost performance of the low resource task of \ac{ttg} and \ac{tth} sign language production.  

%% file: sections/3_method.tex
\section{Methodology}

The task of neural sign language production aims to map a source sequence of spoken language sentences, \(x = (x_{1},x_{2},...,x_{W})\) with W words, to a sequence of glosses, \(y = (y_{1}, y_{2},...,y_{G})\) with G glosses (\acf{ttg}), or a sequence of HamNoSys, \(z = (z_{1}, z_{2},...,z_{H})\) with H symbols (\acf{tth}). \ac{ttg} and \ac{tth} tasks thus learn the conditional probabilities \(p(y|x)\) and \(p(z|x)\) respectively. Sign language translation is not a one to one mapping as several words can be mapped to a single gloss (\(W>G\)), (\(W>H\)). This increases the complexity of the problem as the model must learn to attend to multiple words in the input sequence.

\begin{figure}[!h]
\begin{center}

\includegraphics[width=0.48\textwidth]{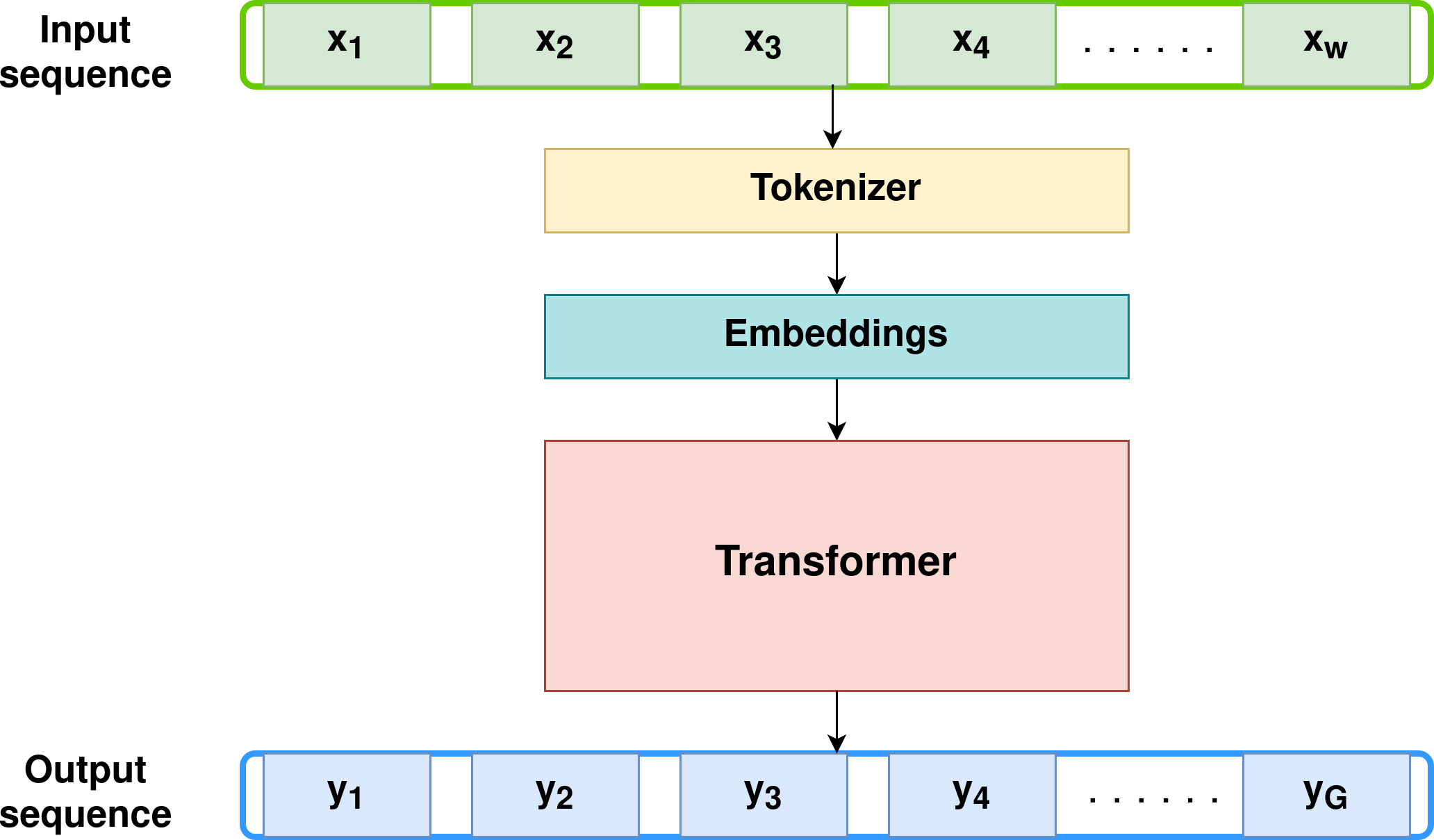} 

\caption{An overview of the different configuration of our architecture for \ac{slt} }
\label{architecture}
\end{center}
\end{figure}

\quad \cref{architecture} shows the general architecture of our model used to translate from spoken language to gloss/HamNoSys. For means of comparison, our baseline model is an encoder-decoder transformer with \ac{mha}. The input and output sequence are tokenized using a word level tokenizer and the embedding for a given sequence is created using a single linear layer. We later build on this base model using different tokenizers, embedding and supervision techniques. We train our model using a cross-entropy loss between the predicted target sequence, \(\hat{x}\) and the ground truth sequence, \({x}^{*}\), defined as \(L_{T}\).    

In this section, we follow the structure of \cref{architecture} from top to bottom. We start by describing the different tokenizers used to split the source text and produce tokens (Sec. 3.1). Next, we explain the different embedding techniques used to create a vector from the input tokens (Sec. 3.2). Finally, we talk about the advantages of using extra supervision and explain how this is implemented in conjunction with the translation loss.   

\subsection{Tokenizers}
Several tokenizer schemes can be used on both the input and output such as \ac{bpe}, Word, character and WordPiece. \ac{bpe} \cite{DBLP:journals/corr/SennrichHB15}, character and WordPiece \cite{schuster2012japanese} all change the vocabulary size of the model by breaking sentences into sub-units. This reduces the number of singletons and reduces lexical inflections in the input and output sequences \cite{wolf2019huggingface}. 

\paragraph{Word} A word level tokenizer segments the input sentence based on white space. Therefore, a normal sentence is split into whole words.  

\paragraph{Character}A character level tokenizer segments the text based on the individual symbols, reducing the vocabulary to simply the alphabet plus punctuation. 

\begin{figure}[!h]
\begin{center}

\includegraphics[width=0.40\textwidth]{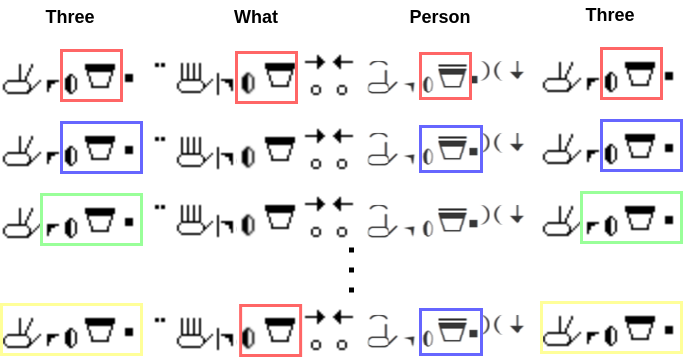} 

\caption{An example of how BPE can be applied to HamNoSys.}
\label{BPE}
\end{center}
\end{figure}

\paragraph{BPE} \ac{bpe} creates a base vocabulary containing all the unique symbols in the data, from which it learns a number of merge rules based on the most commonly occurring sequential symbols. An example of the \ac{bpe} algorithm being applied to HamNoSys is shown in \cref{BPE}, with the coloured boxes indicating what merges are made at each step. Merging continues until a specific vocabulary size is reached. This helps reduce word inflections e.g. the words low, lowest and lower can be segmented to low, est and er. Over the whole corpus the suffix's (est and er) can be reused, collapsing the vocabulary in this example from 3 to 1.

\paragraph{WordPiece}We only apply a WordPiece tokenizer when embedding with BERT, as this is what the BERT model was trained with. WordPiece is another sub-unit tokenization algorithm similar to \ac{bpe} that evaluates the lost benefit before merging two symbols, ensuring that all mergers are beneficial. 

\subsection{Embedding}
After tokenization, the input sequence \(x\) is then embedded by projecting the sequence into a continuous space \cite{mikolov2013efficient}. The goal of embedding is to minimise the Euclidean distance between words with similar meanings. The most common embedding is a single linear layer, which takes an input sequence \(x = (x_{1},x_{2},...,x_{W})\) with W words and turns it into a matrix of [\(W \times E\)] where E is the models embedding width. In models such as BERT and Word2Vec, embeddings are learnt via training on a large corpus of spoken language data. To maximise the benefit from using BERT we fine tune the pre-trained model on the mineDGS dataset using masked-language modeling. 

\quad When using a BERT model, we define the transformation as follows. Given an input sequence \(x\) we first apply WordPiece tokenization. \begin{equation} X_{WP} = WordPiece(x) \end{equation} Then apply the BERT embeddings as: \begin{equation}X_{BERT} = BERT(X_{WP}) \end{equation}Note that we take the embedding from the last layer of BERT. We define the Word2Vec transformation as: \begin{equation} X_{W2V} = Word2Vec(x) \end{equation}

\quad Additionally, we experiment with concatenating or fusing contextual information into the input \(x\). We define the contextual information as \(x_{ave}\) and the scaling factor as \(S\), used to place additional emphasis on the contextual information. In the case of Word2Vec we take a mean average of each word's embedding in the sentence and treat this as a vector that contains information about the whole sentence. For BERT we use the embedding of the classification token ([CLS]), which contains contextual information about the sentence \cite{devlin-etal-2019-bert}. We either concatenate the information to the beginning of a sequence  \(x = (x_{ave}*S,x_{1},x_{2},...,x_{W})\) (CON), or we fuse it into each step of the sequence \(x = ((x_{ave}*S)+x_{1},(x_{ave}*S)+x_{2},...,(x_{ave}*S)+x_{W})\) (ADD). 

\subsection{Supervision}
In sign language, there exists a strong correlation between hand shape and meaning \cite{stokoe1980sign}. Therefore, we investigate forcing the transformer to predict the hand shape alongside the gloss or HamNoSys sequences, to enrich the learnt representation. We scale the loss from the hand shape prediction $\mathcal{L_{H}}$ by factor \(F\). We combine both losses from the translation $\mathcal{L_{T}}$ and hand shape prediction $\mathcal{L_{H}}$ to create \(L_{total}\) as: \begin{equation} L_{total} = L_{T} + (L_{H}*F) \end{equation}
 In this setup, the model learns the joint conditional probability of  \begin{equation} p(y|x)*p(H|x) \end{equation} where \(H\) is the sequence of hand shape symbols: \begin{equation} H = (h_{1},h_{2},...,h_{G}) \end{equation} and G is the number of glosses in the sequence. Overall this forces that model to focus on hand shape during training. We show that by forcing the model to predict hand shape we improve the performance on \ac{tth}.

%% file: sections/4_eperimental.tex
\section{Experiments}

In this section we test the translation performance of our models in both the \ac{ttg} and \ac{tth} setups. We first explain the experimental setup of our models. Next, we compare quantitative results against previous state-of-the-art and our own baselines. Finally we provide qualitative results.  

\subsection{Experimental Setup}
 When training our \ac{ttg} model, we experiment with different embedding sizes, number of layers and heads. We observe a large change in performance based on these three parameters, and search for the best configurations for further tests. Our transformer uses a xavier initializer \cite{glorot2010understanding} with zero bias and Adam optimization \cite{kingma2014adam} with a learning rate of \(10^{-4}\). We also employ dropout connections with a probability of 0.2 \cite{srivastava2014dropout}. When decoding, we use a beam search with a search size of 5.

\quad Our code base comes from \newcite{kreutzer2019joey} NMT toolkit, JoeyNMT \cite{kreutzer2019joey} and is implemented using Pytorch. While our \ac{bpe} and word piece tokenizers come from Huggingface's python library transformers \cite{wolf2019huggingface}. When embedding with BERT, we use an open source pre-trained model from Deepset \cite{chan-etal-2020-germans}. Finally we used fasttext's implementation of Word2Vec for word level embedding \cite{mikolov2013distributed}. 

\quad The publicly available \ac{mdgs} dataset contains aligned spoken German sentences and their gloss counter parts, from unconstrained dialogue between two native deaf signers \cite{kaur2014hamnosys}. The providers of this dataset also have a dictionary for all glosses in the corpus, of which some contain HamNoSys descriptions. Following the translation protocols set in \newcite{saunders2021signing}, we created a subset of the \ac{mdgs} dataset with aligned sentences, glosses and HamNoSys. \ac{mdgs} is a larger dataset compared to \ac{ph14t} (7.5 times more parallel examples, with a source vocabulary of 18,457) with 330 deaf participants performing free form signing. The size of \ac{mdgs} overcomes some of the limitation of PHOENIX2014T. Note we remove the gloss variant numbers to reduce singletons. 

\quad We use the \ac{ph14t} \cite{camgoz2018neural} dataset to compare our best model to previous \ac{nmt} baseline results \cite{saunders2020progressive,stoll2018sign,moryossef2021data,li2021transcribing}. \ac{ph14t} contains parallel monolingual German data, with approximately 7000 examples of aligned gloss and text.

%% file: sections/5_quantitative.tex
\subsection{Quantitative Evaluation} 
 
In this section, we evaluate our models on both \ac{mdgs} and \ac{ph14t} using BLEU (BLEU-1,2,3 and 4) and Rouge (F1-score) scores for both dev and test sets. We group our experiments in five sections:

 \begin{enumerate}
    \itemsep0em 
     \item Baseline \ac{ttg}, \ac{tth} and \ac{ttgth} with a standard transformer. 
     \item T2G and T2H with different embedding layers and sentence averaging. 
     \item T2G and T2H with different tokenizers (BPE, Word, and Character).
     \item T2G and T2H with additional supervision. 
     \item Comparison of our approach on \ac{ph14t} and \ac{mdgs}.
 \end{enumerate}

\input{sections/Tables/embedding}

\input{sections/Tables/tokenizer}

Note we expect the performance to be lower than 100 BLEU. As this is a translation problem there are several valid answers for a given input, thus human evaluation is still necessary. We are also unable to provide \ac{tth} results on \ac{ph14t}, as HamNoSys is not available for some words in its vocabulary.  

\subsubsection{Baseline Results}

Our baseline models achieved a BLEU-4 score of 2.86 (\ac{ttg}), 16.26 (\ac{ttgth}) and 14.46 (\ac{tth}) on the \ac{mdgs} dev set. 
Our baseline setup uses a word level tokenizer on both the input and output, providing a baseline to ablate our proposed techniques in the next three sections. We perform a hyper-parameter search and make modification to the model architecture (number of heads, layers and embedding size) to find the best performance.      

\quad In general, a sequence of HamNoSys is significantly longer than it's gloss counter part, (\(H>>G\)). As a result our \ac{tth} performance is artificially higher than our \ac{ttg}. Therefore, in order to make our T2G and T2H results comparable, we perform a dictionary lookup to convert the gloss to HamNoSys (\ac{ttgth}) before calculating the BLEU score.

Given these results, we conclude a transformer architecture is the best baseline approach and continue with this setup for all future experiments. 

\input{sections/Tables/supervision}

\subsubsection{Embedding}

Next we experiment with using different embedding techniques for the T2G and T2H tasks. As discussed in Section 3.1 we use a linear layer, BERT and Word2Vec in combination with sentence averaging. From the results in \cref{tab:text_to_gloss_embedding_results} we make several observations. Firstly, using a language model improves the translation performance on the \ac{tth} task (Tab. \ref{tab:text_to_gloss_embedding_results}a). While on the \ac{ttg} task, using language models was detrimental to the translation performance (Tab. \ref{tab:text_to_gloss_embedding_results}b). We assume this is due to the reduced information within the gloss and smaller sequence length. Secondly, we observe that applying sentence averaging to the BERT embedding has a negative effect on the scores, independent of what type of average was used (adding or concatenating). On the other hand, adding the sentence averaging to the Word2Vec embedding marginally improved performance compared to the stand alone Word2Vec embeddings on \ac{tth}. But note that Word2Vec plus sentence averaging still has lower performance than just using a linear layer. Overall, we find the best performing embedding to come from using BERT, which scored 5.8 BLEU-4 higher than using a linear layer. This demonstrates that using a pre-trained language model can enhance translation. 

\subsubsection{Tokenizer}

We next experiment with using different tokenizers, as described in Section 3.2. We performed a parameter search to find the best vocabulary size for the BPE algorithm, which we find to be 2250 and 7000 on the input and output respectively. The result of our experiments are shown in \cref{tab:text_gloss_tokenizer_results}. 

\quad When using a character level tokenizer each input contains a minimal amount of information (one letter). As expected this increases the difficulty of the problem, and reduces performance. When applied to the input it was extremely detrimental for the performance on both \ac{ttgth} and \ac{tth}, independent of which output tokenizer was used. Therefore to save space, we do not present the input character level results. Using a word level tokenizer achieved very reasonable results, supporting our theory that using larger units of language that contains more information is beneficial for translation. But as \ac{bpe} outperformed the word level tokenizer on the BLEU-4 score, we assume that by using whole words we create a harder problem, as the dataset contains several word inflections. We conclude that BPE is the best algorithm to use when translating from \ac{tth}. This is due to the algorithms ability to reduce inflections and reduce the vocabulary size which simplifies the networks task. Our results also show that the biggest impact comes from having BPE on the output, suggesting that most of the challenge comes from the decoding section of the network. Similarly, the best \ac{ttg} result came from using a word level and BPE tokenizers on the input and output respectively. 

\subsubsection{Supervision}

Our final ablation study investigates an additional loss explained in Section 3.3. This had a positive effect on the translation performance for \ac{tth}. As can be seen from \cref{tab:translation_with_supervision}, the use of supervision increased the BLEU-4 scores by 0.85. We conclude supervision enriches the learnt sign language representation due to the correlation between hand shape and context. Supervision forces the model to focus more on hand shape, allowing the model to group signs and find better trends in the data. Although the use of supervision marginally decreased the T2G2H BLEU score, we suggest this is due to reduced information in the target gloss. 

\subsection{State-of-the-art Comparisons}

Finally, in \cref{pheonix_baseline} (\ac{ph14t}) and \ref{tab:text_to_gloss_results} (\ac{mdgs}) we compare our best performing models to state-of-the-art work.  
Note in \cref{pheonix_baseline} our baseline is marginally higher than \cite{saunders2020progressive}, we assume this is due to a larger hyper-parameter search. On both datasets, our best model for \ac{ttg} and \ac{ttgth} uses a word level and BPE tokenizer on the input and output respectively. While our best \ac{tth} result comes from adding additional supervision on to this setup. As can be seen from \cref{pheonix_baseline} and \ref{tab:text_to_gloss_results} our models outperformed all other methods  \cite{moryossef2021data,li2021transcribing,saunders2020progressive,saunders2021signing,stoll2018sign}, setting a new state-of-the-art on \ac{ph14t} and \ac{mdgs}. Note we can only compare scores that are publicly available, therefore '-' denotes where the authors did not provide results. 

\input{sections/Tables/PH_baseline}

\input{sections/Tables/mDGS_baseline}

%% file: sections/Tables/embedding.tex
\begin{table*}[!htbp]
\centering
\resizebox{0.75\linewidth}{!}{
\begin{tabular}{@{}p{2.8cm}ccccc|ccccc@{}}
\toprule
            & \multicolumn{5}{c}{DEV SET} & \multicolumn{5}{c}{TEST SET} \\ 
\multicolumn{1}{c|}{Approach:}      & BLEU-4         & BLEU-3         & BLEU-2         & BLEU-1        & ROUGE          & BLEU-4        & BLEU-3         & BLEU-2         & BLEU-1         & ROUGE          \\ \midrule
\multicolumn{1}{r|}{Linear Layer} 
& \textbf{16.26}       & \textbf{24.14}         & \textbf{32.83}        & \textbf{43.05}       & \textbf{42.02}     & \textbf{16.47}      & \textbf{24.51}       & \textbf{33.27}        & \textbf{43.58}        & \textbf{41.53}\\
\multicolumn{1}{r|}{BERT} 
& 14.69 & 21.51 & 29.39  & 38.66  & 30.87  & 14.2 & 21.19 & 29.09  & 38.33  & 30.31      \\
\multicolumn{1}{r|}{BERT SA ADD} 
& 13.23 & 19.41 & 26.43 & 34.75 & 32.38& 13.43 & 19.47 & 26.31 & 34.3 & 32.34          \\
\multicolumn{1}{r|}{BERT SA CON} 
& 14.89 & 21.45 & 28.73 & 36.85 & 34.73& 15.14 & 21.57 & 28.79 & 36.91 & 34.44         \\
\multicolumn{1}{r|}{Word2Vec} 
& 11.47       & 17.59         & 24.68        & 34.21       & 29.45    & 11.73      & 17.83       & 25.14        & 34.90        & 30.22          \\
\multicolumn{1}{r|}{Word2Vec SA ADD} 
& 13.8 & 21.07 & 29.72 & 42.29 & 30.65& 13.31 & 20.56 & 29.31 & 42.13 & 30.67         \\
\multicolumn{1}{r|}{Word2Vec SA CON} 
& 0.03 & 0.05 & 0.06 & 0.04 & 9.44& 0.03 & 0.06 & 0.06 & 0.04 & 9.32         \\
\bottomrule
\end{tabular}
}
\newline \newline
\quad\quad\quad\quad\quad\quad\quad\quad (a) \acf{mdgs} on \acf{ttgth}
\newline

\resizebox{0.75\linewidth}{!}{
\begin{tabular}{@{}p{2.8cm}ccccc|ccccc@{}}
\toprule
            & \multicolumn{5}{c}{DEV SET} & \multicolumn{5}{c}{TEST SET} \\ 
\multicolumn{1}{c|}{Approach:}      & BLEU-4         & BLEU-3         & BLEU-2         & BLEU-1        & ROUGE          & BLEU-4        & BLEU-3         & BLEU-2         & BLEU-1         & ROUGE          \\ \midrule
\multicolumn{1}{r|}{Linear Layer} 
& 14.46       & 23.27         & 32.62        & 47.44       & 50.85    & 14.80      & 23.54       & 32.89        & 47.36        & 50.87          \\
\multicolumn{1}{r|}{BERT} 
& \textbf{20.26}       & \textbf{29.14}         & \textbf{38.01}        & \textbf{48.92}       & \textbf{53.67}    & \textbf{21.03}      & \textbf{29.87}       & \textbf{38.79}        & \textbf{49.77}        & \textbf{53.93}          \\
\multicolumn{1}{r|}{BERT SA ADD} 
& 14.64       & 22.33         & 30.91        & 43.99       & 50.30    & 15.16      & 22.92       & 31.41        & 44.21        & 50.33          \\
\multicolumn{1}{r|}{BERT SA CON} 
& 11.82       & 19.2         & 27.39        & 40.58       & 53.36    & 12.21      & 19.39       & 27.44        & 40.48        & 53.67          \\
\multicolumn{1}{r|}{Word2Vec} 
& 16.43       & 24.77         & 33.71        & 46.62       & 51.14    & 17.09      & 25.23       & 34.22        & 47.31        & 51.52          \\
\multicolumn{1}{r|}{Word2Vec SA ADD} 
& 16.72       & 25.14         & 34.39        & 48.00       & 51.28    & 16.98      & 25.31       & 34.59        & 48.08        & 51.12          \\
\multicolumn{1}{r|}{Word2Vec SA CON} 
& 14.98       & 22.49         & 30.65        & 42.42       & 51.11    & 15.18      & 22.65       & 30.80        & 42.75        & 50.10          \\
\bottomrule
\end{tabular}
}
\newline \newline
(b) \acf{mdgs} on \acf{tth}
\caption{Embedding transformer results for \acf{ttg} and \acf{tth} translation.}
\label{tab:text_to_gloss_embedding_results}
\end{table*}

%% file: sections/Tables/tokenizer.tex
\begin{table*}[!htbp]
\centering
\resizebox{0.75\linewidth}{!}{%
\begin{tabular}{@{}p{2.8cm}c|ccccc|ccccc@{}}
\toprule
\multicolumn{2}{c}{Tokenizer} & \multicolumn{5}{c}{DEV SET}  & \multicolumn{5}{c}{TEST SET} \\ 
 \multicolumn{1}{r}{Input} & Output  & BLEU-4         & BLEU-3         & BLEU-2         & BLEU-1         & ROUGE          & BLEU-4         & BLEU-3         & BLEU-2         & BLEU-1         & ROUGE          \\ \midrule
 \multicolumn{1}{r}{Word}        & Word 
& 16.47 & 24.35 & 33.06 & 43.41 & 36.32 & 16.55 & 24.45 & 33.14 & 43.54 & 36.34          \\
 \multicolumn{1}{r}{Word}        & BPE 
& \textbf{22.06} & \textbf{28.53} & \textbf{36.32} & \textbf{47.55} & 36.20 & \textbf{21.87} & \textbf{28.31} & \textbf{36.02} & \textbf{47.08} & 35.74          \\
 \multicolumn{1}{r}{Word}        & Char 
& 16.47 & 24.35 & 33.06 & 43.41 & 36.32& 16.55 & 24.45 & 33.14 & 43.54 & 36.34          \\
\multicolumn{1}{r}{BPE}        & Word 
& 20.84 & 26.77 & 34.02 & 44.77 & 35.31 & 20.84 & 26.80 & 34.12 & 44.97 & 35.35          \\
\multicolumn{1}{r}{BPE}        & BPE 
& 21.39 & 27.28 & 34.31 & 43.86 & \textbf{36.61}& 21.28 & 27.25 & 34.34 & 43.86 & \textbf{36.86}         \\
\multicolumn{1}{r}{BPE}       & Char 
& 1.99 & 5.5 & 10.35 & 30.01 & 2.61& 1.46 & 5.18 & 10.0 & 29.77 & 2.61          \\
\bottomrule
\end{tabular}%
}
\newline \\
\quad\quad\quad\quad\quad\quad\quad\quad (a) \acf{mdgs} on \acf{ttgth}
\newline

\resizebox{0.75\linewidth}{!}{%
\begin{tabular}{@{}p{2.8cm}c|ccccc|ccccc@{}}
\toprule
\multicolumn{2}{c}{Tokenizer} & \multicolumn{5}{c}{DEV SET}  & \multicolumn{5}{c}{TEST SET} \\ 
\multicolumn{1}{c}{Input} & Output  & BLEU-4         & BLEU-3         & BLEU-2         & BLEU-1         & ROUGE          & BLEU-4         & BLEU-3         & BLEU-2         & BLEU-1         & ROUGE          \\ \midrule
 \multicolumn{1}{r}{Word}       & Word 
& 21.81 & \textbf{31.86} & \textbf{42.05} & \textbf{54.88} & \textbf{55.39}& 21.89 & \textbf{31.92} & \textbf{42.16} & \textbf{55.04} & \textbf{55.23}         \\
 \multicolumn{1}{r}{Word}       & BPE 
& 25.41 & 29.28 & 34.11 & 41.25 & 48.03& 25.54 & 29.39 & 34.25 & 41.35 & 48.09        \\
 \multicolumn{1}{r}{Word}      & Char 
& 21.63 & 31.76 & 41.99 & 54.94 & 55.3& 21.59 & 31.79 & 42.09 & 55.01 & 55.14         \\
\multicolumn{1}{r}{BPE}        & Word 
& 20.98 & 30.29 & 39.38 & 50.04 & 55.24& 21.18 & 30.37 & 39.4 & 49.84 & 55.04        \\
\multicolumn{1}{r}{BPE}        & BPE 
& \textbf{26.14} & 30.83 & 36.47 & 44.35 & 49.95 & \textbf{26.21} & 30.84 & 36.43 & 44.14 & 50.05      \\ 
\multicolumn{1}{r}{BPE}        & Char 
& 1.91 & 6.01 & 11.72 & 37.56 & 37.22& 1.92 & 5.88 & 11.59 & 37.63 & 37.31         \\
\bottomrule
\end{tabular}%
}
\newline \\
(b) \acf{mdgs} on \acf{tth}

\caption{Tokenizer transformer results for \acf{ttg} and \acf{tth} translation.}
\label{tab:text_gloss_tokenizer_results}
\end{table*}

%% file: sections/Tables/supervision.tex
\begin{table*}[ht]
\centering
\resizebox{0.75\linewidth}{!}{%
\begin{tabular}{@{}p{2.8cm}cccccc|ccccc@{}}
\toprule
            & \multicolumn{5}{c}{DEV SET} & \multicolumn{5}{c}{TEST SET} \\ 
\multicolumn{1}{c|}{Approach:} & \multicolumn{1}{c|}{Supervision}      & BLEU-4         & BLEU-3         & BLEU-2         & BLEU-1        & ROUGE          & BLEU-4        & BLEU-3         & BLEU-2         & BLEU-1         & ROUGE          \\ \midrule
\multicolumn{1}{r|}{T2G2H} & \multicolumn{1}{c|}{\xmark} & \textbf{22.06} & \textbf{28.53} & \textbf{36.32} & \textbf{47.55} & \textbf{36.20} & \textbf{21.87} & \textbf{28.31} & \textbf{36.02} & \textbf{47.08} & \textbf{35.74}          \\
\multicolumn{1}{r|}{T2G2H} & \multicolumn{1}{c|}{\cmark} & 21.79 & 27.98 & 35.45 & 46.21 & 35.79 & 21.49 & 27.76 & 35.27 & 46.11 & 35.99             \\
\hline \\
\multicolumn{1}{r|}{T2H} & \multicolumn{1}{c|}{\xmark} & 26.14 & 30.83 & \textbf{36.47} & \textbf{44.35} & \textbf{49.95} & 26.21 & 30.84 & \textbf{36.43} & \textbf{44.14} & \textbf{50.05}       \\ 
\multicolumn{1}{r|}{T2H} & \multicolumn{1}{c|}{\cmark} & \textbf{26.99} & \textbf{31.07} & 35.99 & 42.73 & 48.89& \textbf{27.37} & \textbf{31.42} & 36.3 & 42.92 & 48.85             \\
\bottomrule
\end{tabular}%
}
\caption{HamNoSys hand shape supervision results for \acf{ttgth} and \acf{tth} translation.}
\label{tab:translation_with_supervision}
\end{table*}

%% file: sections/Tables/PH_baseline.tex
\begin{table*}[ht]
\centering
\resizebox{0.75\linewidth}{!}{%
\begin{tabular}{@{}p{2.8cm}ccccc|ccccc@{}}
\toprule
            & \multicolumn{5}{c}{DEV SET} & \multicolumn{5}{c}{TEST SET} \\ 
\multicolumn{1}{c|}{Approach:}      & BLEU-4         & BLEU-3         & BLEU-2         & BLEU-1        & ROUGE          & BLEU-4        & BLEU-3         & BLEU-2         & BLEU-1         & ROUGE          \\ \midrule
\multicolumn{1}{r|}{T2G \cite{stoll2018sign}} & 
16.34       & 22.30         & 32.47        & 50.15       & 48.42    & 15.26      & 21.54       & 32.25        & 50.67        & 48.10          \\
\multicolumn{1}{r|}{T2G \cite{saunders2020progressive}}  &  { 20.23} & { 27.36} & { 38.21} & { 55.65} & { 55.41} & { 19.10} & { 26.24} & { 37.10} & { 55.18} & { 54.55} \\ 
\multicolumn{1}{r|}{T2G \cite{li2021transcribing}} & 
18.89       & 25.51         & -        & -       & 49.91    & -      & -       & -        & -        & -          \\
\multicolumn{1}{r|}{T2G \cite{moryossef2021data}} & 
23.17       & -         & -        & -       & -    & -      & -       & -        & -        & -          \\
\multicolumn{1}{r|}{} & 
       &        &         &        &    &      &        &         &       &  \\
\multicolumn{1}{r|}{\textbf{T2G Baseline (ours)}} & 
22.47       & 30.03         & 41.54        & 58.98       & 57.96    & 20.95      & 28.50       & 39.99        & 58.32        & 57.28          \\
\multicolumn{1}{r|}{\textbf{T2G Best Model (ours)}} & 
\textbf{25.09} & \textbf{32.18} & \textbf{42.85} & \textbf{60.04} & \textbf{58.82} & \textbf{23.19} & \textbf{30.24} & \textbf{40.86} & \textbf{58.74} & \textbf{56.55} \\
\bottomrule
\end{tabular}%
}
\caption{Baseline comparison results for \acf{ttg} translation on \ac{ph14t}.}
\label{pheonix_baseline}
\end{table*}

%% file: sections/Tables/mDGS_baseline.tex
\begin{table*}[ht]
\centering
\resizebox{0.75\linewidth}{!}{%
\begin{tabular}{@{}p{2.8cm}ccccc|ccccc@{}}
\toprule
            & \multicolumn{5}{c}{DEV SET} & \multicolumn{5}{c}{TEST SET} \\ 
\multicolumn{1}{c|}{Approach:}      & BLEU-4         & BLEU-3         & BLEU-2         & BLEU-1        & ROUGE          & BLEU-4        & BLEU-3         & BLEU-2         & BLEU-1         & ROUGE          \\ \midrule
\multicolumn{1}{r|}{T2G \cite{saunders2021signing}} & 
3.17       & -         & -        & -       & 32.93     & 3.08      & -       & -        & -        & 32.52          \\ 
\multicolumn{1}{r|}{} & 
       &        &         &        &    &      &        &         &       &  \\
\multicolumn{1}{r|}{\textbf{T2G Our best}} 
& 10.5 & 14.35 & 20.43 & 33.56 & 35.79& 10.4 & 14.21 & 20.2 & 33.59 & 35.99          \\
\multicolumn{1}{r|}{\textbf{T2G2H Our best}} 
& 22.06 & 28.53 & 36.32 & 47.55 & 36.20 & 21.87 & 28.31 & 36.02 & 47.08 & 35.74         \\
\multicolumn{1}{r|}{\textbf{T2H Our best}} 
& \textbf{26.99} & \textbf{31.07} & \textbf{35.99} & \textbf{42.73} & \textbf{48.89} & \textbf{27.37} & \textbf{31.42} & \textbf{36.3} & \textbf{42.92} & \textbf{48.85}   \\
\bottomrule
\end{tabular}%
}
\caption{Baseline comparison results for \acf{ttg}, \acf{ttgth} and \acf{tth} translation on \ac{mdgs}.}
\label{tab:text_to_gloss_results}
\end{table*}

%% file: sections/6_qualitative.tex
\subsection{Qualitative Evaluation}
\begin{figure}[!h]
\begin{center}

\includegraphics[width=0.30\textwidth]{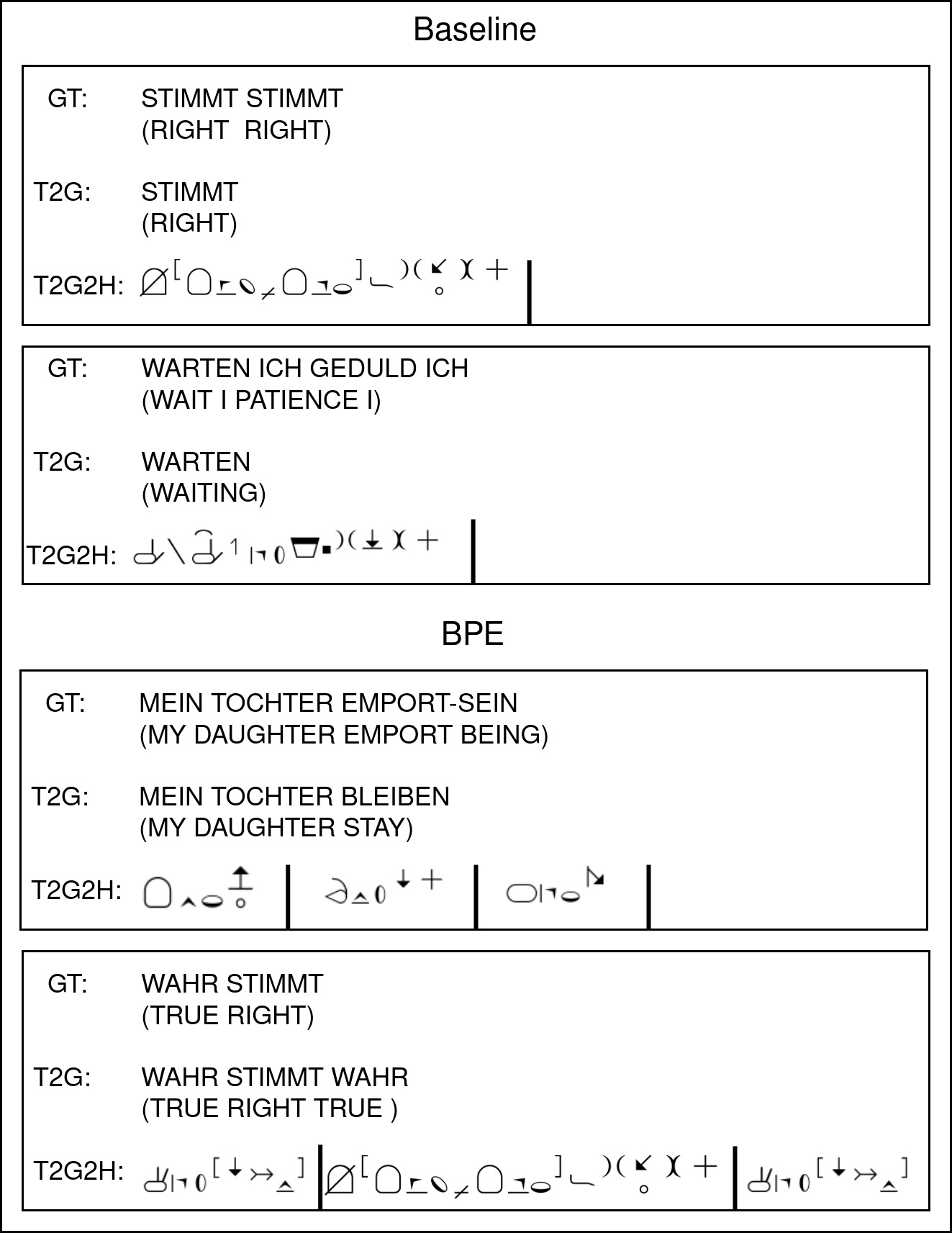} 

\caption{Translation examples from our baseline and best model.}
\label{translation_exmples}
\end{center}
\end{figure}

For qualitative evaluation, we share translation examples from our best models and our baseline model in \cref{translation_exmples}, to allow the reader to better interpret the results. Note, we add a vertical black line after each word of HamNoSys to mark the end of a given sign. These results show how our BPE model has learnt richer translations than our baseline model.   

